\documentclass[journal]{IEEEtran}
\usepackage{cite}
\usepackage{amsmath,amssymb,amsfonts}
\usepackage{algorithmic}
\usepackage{graphicx}
\usepackage{textcomp}
\usepackage{wrapfig}
\usepackage{moreverb,url}
\usepackage{graphicx}
\usepackage{subfigure}
\usepackage{amsmath} 
\usepackage{amsmath,amssymb,amsfonts}
\usepackage{algorithmic}
\usepackage[linesnumbered,ruled]{algorithm2e}
\SetKwBlock{Predict}{\underline{Predict:}}{end\ Predict}
\SetKwBlock{Correction}{\underline{Correction:}}{end\ Correction}
\usepackage{graphicx}
\usepackage{textcomp}
\usepackage[table, svgnames, dvipsnames]{xcolor}
\usepackage{tikz}
\usetikzlibrary{calc,patterns,angles,quotes}
\usetikzlibrary{arrows,automata,positioning,decorations.markings,calc,shapes}
\usetikzlibrary{datavisualization.formats.functions}
\usetikzlibrary{decorations.pathreplacing,calligraphy}
\usetikzlibrary{intersections}
\usepackage{orcidlink}
\usepackage{hyperref}
\hypersetup{colorlinks=true, linkcolor=blue, citecolor=green, urlcolor=blue, bookmarksdepth=3}
\begin{document}

\title{Enhanced Unscented Kalman Filter-Based SLAM in Dynamic Environments: Euclidean Approach}

\author{Masoud Dorvash~\orcidlink{0009-0003-9463-9031}, Ali Eslamian~\orcidlink{0000-0001-9754-9865}, and Mohammad Reza Ahmadzadeh~\orcidlink{0000-0001-9558-5854}}



\maketitle

\begin{abstract}
This paper introduces an innovative approach to Simultaneous Localization and Mapping (SLAM) using the Unscented Kalman Filter (UKF) in a dynamic environment. The UKF is proven to be a robust estimator and demonstrates lower sensitivity to sensor data errors compared to alternative SLAM algorithms. However, conventional algorithms are primarily concerned with stationary landmarks, which might prevent localization in dynamic environments. This paper proposes an Euclidean-based method for handling moving landmarks, calculating and estimating distances between the robot and each moving landmark, and addressing sensor measurement conflicts. The approach is evaluated through simulations in MATLAB and comparing results with the conventional UKF-SLAM algorithm. We also introduce a dataset for filter-based algorithms in dynamic environments, which can be used as a benchmark for evaluating of future algorithms. The outcomes of the proposed algorithm underscore that this simple yet effective approach mitigates the disruptive impact of moving landmarks, as evidenced by a thorough examination involving parameters such as the number of moving and stationary landmarks, waypoints, and computational efficiency. We also evaluated our algorithms in a realistic simulation of a real-world mapping task. This approach allowed us to assess our methods in practical conditions and gain insights for future enhancements. Our algorithm surpassed the performance of all competing methods in the evaluation, showcasing its ability to excel in real-world mapping scenarios.
\end{abstract}

\begin{IEEEkeywords}
Autonomous Navigation, Dynamic Environments, Filter-Based SLAM, Robotic , Simultaneous Localization and Mapping, Unscented Kalman Filter-SLAM
\end{IEEEkeywords}

\section{Introduction}
\IEEEPARstart{S}{imultaneous} Localization and Mapping (SLAM) is a foundational technology in robotics, serving as the basis for a wide range of applications in both academic research and industry \cite{van2021solid}. The emergence of enhanced computer processing capabilities and a growing number of inexpensive sensors, such as cameras and laser rangefinders, has increased the scope of SLAM's practical application over time. This expansion includes a wide range of scenarios, such as guiding domestic robotic vacuums and coordinating the movements of autonomous robots in warehouses, as well as facilitating autonomous parking and increasing drone deliveries across diverse and challenging landscapes \cite{taketomi2017visual}.

SLAM algorithms are determined by the characteristics of their inputs, which may include images or sensor-derived range data, which result in visual-based and filter-based algorithms \cite{muhammad2009current}. In the past, these algorithms have been optimized for indoor or restricted surroundings, where the environmental effects are typically assumed to be stationary. However, the actual world frequently challenges this presumption and presents dynamic complexities that necessitate innovative approaches. As a result, various strategies to confront this challenge head-on have emerged \cite{10043931}. While deep learning techniques are usually associated with image inputs, the utility of geometric constraints is not limited to visual data alone \cite{li2022overview}. Instead, these constraints can be effectively exploited by filter-based approaches, which incorporate more comprehensive range of data types, including sensor-derived data. This flexibility in applying of geometric constraints highlights their importance in addressing the complexities of dynamic environments across different data types.
\IEEEpubidadjcol

\subsection{Contributions}
In this paper, we explore a detailed and comprehensive formulation of the Unscented Kalman Filter (UKF) SLAM framework. Our method determines the movement of landmarks by comparing the Euclidean distance between the robot and a landmark at a given step with the estimated distance for the subsequent step, using the robot's own compass. This foundational understanding leads us to introduce a novel, low-complexity algorithm based on this distance comparison, capable at differentiating between stationary and moving landmarks. Leveraging this technique to filter out moving landmarks from observations, we observe a marked improvement in the accuracy of robot path prediction and movement. A distinctive feature of our method is its inherent simplicity and low computational complexity, ensuring it is both practical and amenable to various real-world applications. 

\subsection{Organization and Notations}
The remainder of the paper is organized as follows: In Section \ref{sec:prev}, we review previous work on SLAM, with a particular focus on UKF-SLAM. In Section \ref{sec:prob}, we introduce the system model for UKF-SLAM. Section \ref{sec:myth} delves into the interaction between UKF and SLAM in dense dynamic environments, and we propose a novel algorithm designed explicitly for this setting. In Section \ref{sec:sim}, we analyze the performance of the newly proposed methods using illustrative examples. Finally, Section \ref{sec:con} presents our conclusions and suggests potential approaches for future research.

In this paper, we employ a consistent notation to enhance clarity. Key notational elements include:
\begin{itemize}
	\item \textbf{Vectors and Matrices}: Vectors are represented in boldface, such as $\boldsymbol{a}$, and matrices are denoted in uppercase boldface, such as $\boldsymbol{A}$. The transpose operation is indicated by $(\cdot)^T$.
	\item \textbf{Real Number Sets}: We use $\mathbb{R}^N$ to refer to the set of $N$-dimensional vectors containing real numbers and $\mathbb{R}^{M\times N}$ for matrices of real numbers with dimensions $M\times N$.
	\item \textbf{Prediction}: The term $\boldsymbol{x}_{k+1|k}$ represents the prediction of the vector $\boldsymbol{x}$ at step $k+1$, considering all available information up to step $k$.
	\item \textbf{Estimation Symbol}: We use vector $\boldsymbol{\hat{x}}$ to represent estimations of the vector $\boldsymbol{x}$.
	\item \textbf{Euclidean Distance}: The Euclidean distance is denoted as $\left\lvert \cdot \right\rvert$.
	\item \textbf{Indexing and Sequence Notation}: The variable $k$ denotes a step within a procedure, while superscripts $i$ and $j$ refer to the $i$-th and $j$-th elements of their respective sets.
\end{itemize}

\section{Previous Works}\label{sec:prev}

\subsection{Static Environment}
The Extended Kalman Filter (EKF) has been a popular choice among SLAM algorithms for a long time, frequently used to estimate both the robot's pose and the locations of observed landmarks. However, EKF-based SLAM is vulnerable to its natural difficulties of linearization errors, which can result in inadequate performance or even divergence. One critical issue with EKF-based SLAM is the tendency for state estimates to become inconsistent which undermines the reliability of the estimator. There has been a notable evolution toward employing the UKF to mitigate the disadvantages of linearization. The UKF, which is well-known for its superior performance in nonlinear estimation problems, has the potential to serve as a compelling alternative for SLAM applications \cite{kurt2012comparison}.

The FastSLAM algorithm \cite{10.5555/1630659.1630824}, estimates the robot's position and the environment map using particle filters. FastSLAM, unlike UKF-SLAM, maintains a set of particles, each of which represents a possible hypothesis regarding the robot's pose and the map. Each particle has its conception of the robot's and landmarks' positions, represented as Gaussian. To update the particles, FastSLAM employs a two-step procedure: first, it updates the robot's pose using the motion model, and then updates the landmark positions using the sensor measurements \cite{huang2009complexity}. This strategy enables FastSLAM to deal with nonlinearities in the system model and the data association problem more efficiently than UKF-SLAM. However, FastSLAM has a greater computational complexity than UKF-SLAM, making it more challenging to implement in real-time applications.

The Adaptive Unscented Kalman Filter (AUKF) algorithm presented in \cite{bozorg2019new} aims to enhance the navigation accuracy of autonomous vehicles. This algorithm provides an improved version of the UKF that continually adjusts filter parameters based on the prevailing system state. The paper thoroughly explains the AUKF algorithm, comparing its performance with the standard UKF and EKF algorithms using simulated and real-world datasets. In a real-world scenario with a two-wheel differential mobile robot navigating an indoor environment with various objects, the results highlight that the AUKF-SLAM algorithm demonstrates superior accuracy and noise robustness.

In reference \cite{bahraini2020efficiency} pioneered a more efficient approach to posterior probability estimation using AUKF and a randomized search technique. Their algorithm dynamically fine-tunes the scaling parameter to optimize posterior probability estimation and constrains computation by the means and covariances of currently observed landmarks. This reduces computation time by 50 percent compared to other state-of-the-art methods, making it suitable for real-time applications. 

\subsection{Dynamic Environment}
Reference \cite{vidal2015slam} presents a novel method for addressing the SLAM problem in dynamic environments, emphasizing robot self-localization using an outliers filter and a modified particle filter. The outliers filter distinguishes between stationary and moving environmental landmarks, discarding the latter to guarantee the accuracy of the SLAM algorithm. By analyzing the correlation between landmark positions, the filter identifies outliers and reduces their impact. This approach not only addresses cases of close feature descriptors but also effectively manages moving landmarks, ultimately enhancing SLAM accuracy. The paper also employs a modified particle filter that estimates the robot's pose novelly, along with continuous resampling of particles based on observations from an RGB-Depth camera in real-time.

The approach presented in \cite{todoran2016extended} utilizes laser range scanning to generate a sparse representation of clustered scan points that dynamically forms the basis for decision-making. This method distinguishes itself from conventional SLAM techniques by employing low-constraint features and not requiring high update rates. It excels at managing dynamic environments by modeling and grouping them as fixed-moving objects, considering processes like object discovery, timeouts, mergers, splits, and symmetries. The algorithm leverages moving object state information to self-correct the agent's position and locally generate maps. It provides categorized objects and their dynamic descriptors, facilitating self-localization, mapping, path-planning, sensor fusion, and various robotic tasks.

The approach in \cite{lee2013slam} models the SLAM estimation problem as a single-cluster process, with vehicle motion as the parent nodes and map features as the daughter nodes. This framework yields tractable recursive formulas for Bayesian filtering. A distinguishing characteristic of the proposed filter is its provision of a robust multi-object likelihood that aligns with the underlying assumptions. The authors employ a Gaussian mixture methodology capable of addressing SLAM challenges such as variable sensor fields of view and a combination of stationary and moving map features to implement the filter. Comprehensive Monte Carlo simulations demonstrate the filter's effectiveness in scenarios with high measurement clutter and nonlinear vehicle motion, validating its potential utility for dynamic SLAM applications.

\section{Problem Formulation}\label{sec:prob}
In this section, we delve into the basic concepts of UKF-SLAM, a crucial component of our research. UKF-SLAM is a recognized robotics method for simultaneous environment perception and robot localization. It maintains a belief distribution over the robot's pose and map, updating it with motion and sensor data. The motion model is used to predict the robot's pose from its previous pose, and the sensor model is used to update the belief distribution based on the sensor measurements.

\subsection{UKF-SLAM in Static Environment}\label{AA}
We will investigate the UKF equations, which are the essential mathematical instrument underlying this technique. The UKF is a nonlinear filter, so it uses a different approach to propagate the belief distribution than a linear filter like the Kalman filter \cite{Candy2016}. The EKF uses a first-order Taylor expansion to linearize nonlinear models. The UKF instead employs the Unscented Transformation \cite{julier2002scaled}. It employs statistical linearization methods to linearize motion and observation equations, facilitating the estimation of both robot and landmark positions.

The state vector $\boldsymbol{x}_{k}$ contains the current position of the robot's sensor including the position of the landmarks in $k$-th step and the related covariance matrix $\boldsymbol{\Sigma}_{k}$ written as:
\begin{equation}\label{eq:muSigSta}
	\begin{split}
		\boldsymbol{x}_{k} &= \begin{bmatrix}
			x_k & y_k & \theta_k & m_{1,x}^s & m_{1,y}^s & \cdots & m_{n,x}^s & m_{n,y}^s
		\end{bmatrix}^T \\
		\boldsymbol{\Sigma}_{k} &= \begin{bmatrix}
			\boldsymbol{\Sigma}_{x_kx_k} & \boldsymbol{\Sigma}_{x_km^s} \\
			\boldsymbol{\Sigma}_{m^sx_k} & \boldsymbol{\Sigma}_{m^sm^s}
		\end{bmatrix}
	\end{split}
\end{equation}
where $x_k$ and $ y_k$ denote the Cartesian position of the robot, while ${\theta}_k$ represents its rotation in the $k$-th step and also 
$m_{n,x}^s$  $m_{n,y}^s$    
are position of the $n$-th landmark (the $s$ superscript indicates the stationary landmark). Let the $N$ represents the size of the state vector; therefore $\boldsymbol{x}_k \in \mathbb{R}^N$ and $\boldsymbol{\Sigma}_k \in \mathbb{R}^{N \times N}$. The system model is defined as follows:
\begin{equation}\label{eq:sysmodel}
	\boldsymbol{x}_{k+1} = f({x}_{k}) =  \begin{bmatrix}
		v\cos(\theta)\cdot dt \\
		v\sin(\theta)\cdot dt \\
		\omega\cdot dt \\
		\boldsymbol{m}
	\end{bmatrix} = \begin{bmatrix}
		x_{k+1} \\
		y_{k+1} \\
		\theta_{k+1} \\
		\boldsymbol{m}
	\end{bmatrix}
\end{equation}
herein $v$ represents the velocity, $\omega$ represents the rotational speed of the robot, $dt$ represents the time interval for each step, and $\boldsymbol{m}$ indicates the map features.

If the size of the state vector is $N$, the UKF employs $2N+1$ weighted and distributed sigma points, which are carefully chosen points that represent the mean and covariance of the Gaussian distribution of the state vector of the robot. The sigma points can be defined as \cite{simon2006optimal}:
\begin{equation}\label{eq:sigmapoints}
	\begin{split}
		\boldsymbol{\chi}_{k|k}^0 &= \hat{\boldsymbol{x}}_{k|k} \\
		\boldsymbol{\chi}_{k|k}^i &= \hat{\boldsymbol{x}}_{k|k} + \left[ \sqrt{(N + \lambda) \boldsymbol{\Sigma}_{k|k}} \right]_i i = 1,2,\ldots,N  \\
		\boldsymbol{\chi}_{k|k}^{i+N} &= \hat{\boldsymbol{x}}_{k|k} - \left[ \sqrt{(N + \lambda) \boldsymbol{\Sigma}_{k|k}} \right]_i i = N+1,\ldots,2N  
	\end{split}
\end{equation}

In \eqref{eq:sigmapoints}, $\lambda$ represents the scaling parameter of the sigma points distribution. Additionally, $\left[ \sqrt{(N + \lambda) \boldsymbol{\Sigma}_{k|k}} \right]_i$ denotes the $i$-th column of the matrix $\left[ \sqrt{(N + \lambda) \boldsymbol{\Sigma}_{k|k}} \right]$. The term $\sqrt{\boldsymbol{\Sigma}_{k|k}}$ refers to the Cholesky decomposition matrix of $\boldsymbol{\Sigma}_{k|k}$. So, the sigma points matrix can be defined as \cite{simon2006optimal}:
\begin{equation}\label{eq:sigmapointsmat}
	\boldsymbol{\chi}_{k|k} = \begin{bmatrix}
		\boldsymbol{\chi}_{k|k}^0 & \ldots & \boldsymbol{\chi}_{k|k}^i & \ldots & \boldsymbol{\chi}_{k|k}^{2N}
	\end{bmatrix}_ {N \times (2N+1)}
\end{equation}
Each sigma point vector $\boldsymbol{\chi}_{k|k}^i$ is weighted using the scaling parameter $\lambda$ according to the following relations \cite{simon2006optimal}.
\begin{equation} \label{eq:weigh}
	\begin{aligned}
		w^0 &= \frac{\lambda}{N + \lambda} & i &= 0 \\
		w^i &= \frac{\lambda}{2(N + \lambda)} & i &= 1, \ldots, 2N
	\end{aligned}
\end{equation}

Furthermore, the transformed sigma points can be expressed using the system model as:
\begin{equation}
	\boldsymbol{\chi}_{k+1|k} = f(\boldsymbol{\chi}_{k|k})
\end{equation}

Following the prediction step, the state vector $\boldsymbol{x}_{k}$ and covariance matrix $\boldsymbol{\Sigma}_{k}$ can be resulted as \cite{simon2006optimal}:

\begin{equation}\label{prediction}
	\begin{aligned}
		&\hat{\boldsymbol{x}}_{k+1|k} = \sum_{i=0}^{2N} w^i \boldsymbol{\chi}_{k+1|k}^i \\
		&\hat{\boldsymbol{\Sigma}}_{k+1|k} = \sum_{i=0}^{2N} w^i \left( \boldsymbol{\chi}_{k+1|k} - \hat{\boldsymbol{x}}_{k+1|k}\right) \left( \boldsymbol{\chi}_{k+1|k} - \hat{\boldsymbol{x}}_{k+1|k}\right)^T
	\end{aligned}
\end{equation}

After the prediction step, the robot's sensors observe the environment and search for landmarks. It should be noted that the positions of the observed landmarks are assumed to be known. However, the state vector will be updated based on the current position of the robot. We define the observed state vector as $\boldsymbol{z}_{k|k-1}$ and its corresponding sigma points as $\boldsymbol{Z}_{k|k-1}$, which are obtained from \cite{simon2006optimal}:
\begin{equation}\label{Zsig}
	\boldsymbol{Z}_{k|k-1}^{i} = g(\boldsymbol{\chi}_{k|k-1}^i), \quad i = 0, \ldots, 2N
\end{equation}
Which $g(.)$ is the observation system model. So, the estimated observed state vector can be written as:
\begin{equation}\label{ztrue}
	\hat{\boldsymbol{z}}_{k|k-1} = \sum_{i=0}^{2N} w^i \boldsymbol{Z}_{k|k-1}^i
\end{equation}

In the correction step, for calculating the Kalman gain matrix, we need the observed covariance matrix $(\boldsymbol{\Sigma}_{zz})$ and also the cross-covariance matrix of state and observed vector $(\boldsymbol{\Sigma}_{xz})$ as follows:
\begin{equation}\label{mutualcovariance}
	\begin{alignedat}{2}
		&\boldsymbol{\Sigma}_{xz,k|k-1} = \sum_{i=0}^{2N}&&\biggl\{ w^i \left( \boldsymbol{Z}_{k|k-1}^i - \hat{\boldsymbol{z}}_{k|k-1}^i\right)\biggr.\\
		& &&\biggl.\left( \boldsymbol{Z}_{k|k-1}^i - \hat{\boldsymbol{z}}_{k|k-1}^i\right)^T\biggr\}  \\ 
		&\boldsymbol{\Sigma}_{xx,k|k-1} = \sum_{i=0}^{2N}&& \biggl\{w^i \left( \boldsymbol{\chi}_{k|k-1}^i - \hat{\boldsymbol{x}}_{k|k-1}^i\right)\biggr.\\
		& &&\biggl. \left( \boldsymbol{Z}_{k|k-1}^i - \hat{\boldsymbol{z}}_{k|k-1}^i\right)^T \biggr\}
	\end{alignedat} 
\end{equation}
So the Kalman gain matrix is:
\begin{equation}\label{eq:kalmangain}
	\boldsymbol{L}_k = \boldsymbol{\Sigma}_{xz,k|k-1} \left( \boldsymbol{\Sigma}_{xx,k|k-1} \right)^{-1}
\end{equation}
The state vector $\hat{\boldsymbol{x}}_{k|k-1}$ and the covariance matrix $\hat{\boldsymbol{\Sigma}}_{k|k-1}$ will be updated using the Kalman gain matrix as follows:
\begin{equation}\label{xest}
	\begin{alignedat}{2}
		&\hat{\boldsymbol{x}}_{k|k} &&= \hat{\boldsymbol{x}}_{k|k-1} + \boldsymbol{L}_k \left( \boldsymbol{z}_k - \hat{\boldsymbol{z}}_{k|k-1} \right)  \\
		&\hat{\boldsymbol{\Sigma}}_{k|k} &&= \hat{\boldsymbol{\Sigma}}_{k|k-1} + \boldsymbol{L}_k \hat{\boldsymbol{\Sigma}}_{zz,k|k-1}\boldsymbol{L}_k^T
	\end{alignedat} 
\end{equation}

The procedure for implementing UKF-SLAM in a static environment is comprehensively outlined, as demonstrated in Algorithm 1.
\begin{algorithm}
	\caption{UKF-SLAM}\label{alg:UKFSLAM}
	\textbf{Initialization}: Initial state vector $\boldsymbol{\mu}_{0} = \begin{bmatrix}x_{0,k}& y_{0,k}& \theta_{0,k}\end{bmatrix}^T$ and initial covariance matrix $\boldsymbol{\Sigma}_{0} = \mathrm{diag}\left(\sigma^2_{x_k} ,\sigma^2_{y_k}, \sigma^2_{\theta_k}\right)$\\
	\For {each step $k$}{
		\Predict{
			Calculate the sigma points \eqref{eq:sigmapoints} and weights \eqref{eq:weigh} \\
			Predict state vector and covariance matrix using \eqref{prediction}
		}
		\Correction{
			Calculate the sigma point of observed vector \eqref{Zsig} \\
			Calculate the observed vector \eqref{ztrue} \\
			Calculate the cross covaiance matrix $\boldsymbol{\Sigma}_{xz}$ and $\boldsymbol{\Sigma}_{zz}$ \eqref{mutualcovariance} \\
			Calculate the Kalman gain matrix \eqref{eq:kalmangain} \\
			Estimate the state vector and covariance matrix \eqref{xest}
		}
	}
\end{algorithm}
\subsection{UKF-SLAM in Dynamic Environment}\label{DyanamicEnv}
In dynamic environments, the primary challenge faced by SLAM algorithms is ensuring the accuracy and consistency of the constructed map \cite{8593691}. Key to this challenge is the necessity to distinguish between stationary and moving landmarks, with a particular emphasis on the latter.

A foundational principle of any map-building effort is its dedication to the consistent representation of an environment. Landmarks traditionally serve as fixed points of reference instrumental in facilitating the localization process. However, moving landmarks, due to their variable nature, can introduce significant inconsistencies in the map \cite{9197349}. Their constantly changing positions in the environment not only reduce the map's reliability but also lead to inaccurate robot pose estimations, subsequently complicating the entire navigation process \cite{Wang2022}.

The efficiency in which the SLAM operates is naturally connected to its computational and memory demands \cite{7001621}. Every landmark incorporated into the system amplifies the complexity of the UKF mechanism, thereby increases its computational consumption \cite{5152793}. Persistently accounting for moving landmarks, despite their temporary nature, challenges this computational system without offering substantial returns in localization precision. At the same time, robots, particularly those reliant on embedded systems, struggle with finite memory resources \cite{9668630}. Allocating memory to unstable information from such landmarks can lead to sub-optimal memory management, potentially impeding other critical robot functions \cite{s22186903}.

Data association, correlating newly observed landmarks with pre-existing ones on the map, constitutes another pivotal aspect of SLAM \cite{9210559}. Moving landmarks, with their changeable positions, introduce a layer of ambiguity to this process \cite{yin2022dynam}. Effectively, their presence complicates identifying a precise match between current observations and the existing map data, potentially leading to data mismatches or inaccuracies \cite{gil2006improving}.

Uncertainty remains an essential attribute of SLAM, spreading through both robot pose and landmark position estimations. Moving landmarks amplify this natural uncertainty. Their unpredictable movements and characteristics introduce many of variables that decrease the accuracy of estimates \cite{xie2020moving}. By actively addressing these landmarks, either through removal or advanced management techniques, this uncertainty can be significantly mitigated.

For extended robotic operations or tasks that necessitate multi-session mapping, the temporal consistency of maps becomes an essential need. Moving landmarks, if unaddressed, can lead to temporal differences in the map, complicating efforts such as revisiting specific regions or merging maps across different sessions \cite{bailey2006simultaneous}. Ensuring that the map primarily reflects the stationary features of the environment ensures its continuous relevance across a diverse range of robotic tasks, from detailed planned moves to complex environment interpretations \cite{liang2021semi}.

While moving landmarks certainly capture the evolving details of an environment, their unchecked presence in UKF-SLAM systems poses substantive challenges. Addressing these landmarks is not just strategically advantageous—it's an imperative that underpins the efficacy and reliability of SLAM in dynamic contexts.

\section{Methodology}\label{sec:myth}
Recognizing the importance of identifying moving landmarks in SLAM, it is crucial to note that specifying a moving object in any setting requires a minimum of two snapshots to decisively determine its motion. Accordingly, our approach also necessitates at least two snapshots for practical evaluation. Our approach leverages the principle of the Euclidean distance property.

Let's consider a scenario where a robot, at step $k$, detects a landmark $z^i_k$. The robot measures the distance between its current position and landmark $z^i_k$. Advancing to the next step, $k+1$, and given our knowledge of the distance passed during each step and the angle between the path traveled and the distance from the robot's position at step $k$ to landmark $z^i_k$, we can estimate the distance between the robot's new position at $k+1$ and landmark $z^i_{k+1}$. For a stationary landmark $z^i$, the newly measured distance and our estimate should ideally equal, but in real application we may use:
\begin{equation}
	\left\lvert \hat{d}^{r,i}_{k+1}- d^{r,i}_{k+1}\right\rvert \leq \varepsilon
\end{equation}
where $\hat{d}^{r,i}_{k+1}$ and $d^{r,i}_{k+1}$ denote the estimated and measured distances, respectively, between the robot at step $k+1$ and the landmark $i$. The parameter $\varepsilon$ is employed as a threshold to determine whether landmark $i$ is in motion. If the inequality holds true, we identify that the landmark remains stationary and subsequently incorporate its properties into the state vector. Otherwise, we conclude that the landmark is moving and exclude it from the state vector.

\begin{figure}
	\centering
	\begin{tikzpicture}[every text node part/.style={align=center},>=latex,node distance=1cm]
		\node[isosceles triangle,isosceles triangle apex angle=40,rotate=45,draw,fill=orange!60,
		minimum size =2mm,thick] at (0,0) (robotK0) {};
		\node[isosceles triangle,isosceles triangle apex angle=40,rotate=15,draw,fill=orange!60,
		minimum size =2mm,thick] at (3,1) (robotK1) {};
		\draw [dashed] (robotK0.east) -- node [below] {$d_s$} (robotK1.west);
		\draw [dashed] (robotK0.east) -- node [below left] {$d_k^{r,i}$} (1,4) node [circle,draw,solid,minimum width=2mm,fill=cyan!60,thick] (land1) {} node [xshift=-0.1cm,left] {$z^i_k$};
		\draw [thick] (land1) -- node [above right] {$\hat{d}_{k+1}^{r,i}$} (robotK1.east);
		\draw [thick] (robotK1.east) -- node [right] {$d_{k+1}^{r,i}$} (3.5,5.5) node [circle,draw,solid,minimum width=2mm,fill=cyan!60,thick] (land2) {} node [xshift=0.1cm,right] {$z^i_{k+1}$};
		\draw [->,thick,color=ForestGreen] (land1) -- (land2);
		\coordinate (robotK0east) at (robotK0.east);
		\coordinate (robotK1east) at (robotK1.east);
		\pic [draw, "$\alpha_{k+1}^i$", angle eccentricity=1.8,angle radius=1cm,angle radius=0.5cm] {angle = robotK1east--robotK0east--land1};
		\node at (robotK0) [below,yshift=-0.3cm] {$r_k$};
		\node at (robotK1) [below,yshift=-0.3cm] {$r_{k+1}$};
	\end{tikzpicture}
	\caption{Distances between the robot and a moving landmark at steps $k$ and $k+1$, highlighting landmark motion.}
	\label{fig:Proposed}
\end{figure}
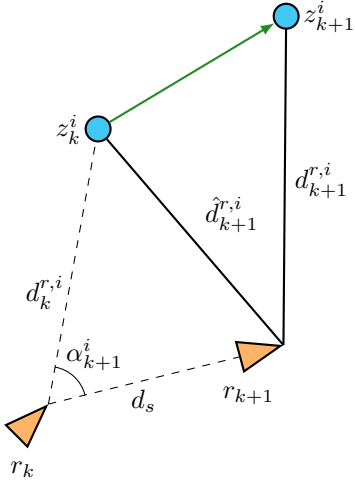

Figure \ref{fig:Proposed} provides a visual representation to elucidate this process. In this figure, the positions of the robot at two successive steps, $k$ and $k+1$, are depicted concerning a moving landmark $z^i$. As observed, the estimated distance $\hat{d}^{r,i}_{k+1}$ does not match the measured distance from the robot at step $k+1$ to landmark $z^i_{k+1}$, leading us to conclude the landmark's motion.

\begin{figure}
	\centering
	\begin{tikzpicture}[every text node part/.style={align=center},>=latex,node distance=1cm]
		\node[isosceles triangle,isosceles triangle apex angle=40,rotate=45,draw,fill=orange!60,
		minimum size =2mm,thick] at (0,0) (robotK0) {};
		\node[isosceles triangle,isosceles triangle apex angle=40,rotate=15,draw,fill=orange!60,
		minimum size =2mm,thick] at (3,1) (robotK1) {};
		\draw [<-,thick,color=ForestGreen] ([shift={(robotK1.east)}]51.5:3.85) arc (51.5:130:3.85);
		\draw [dashed] (robotK0.east) -- node [below] {$d_s$} (robotK1.west);
		\draw [dashed] (robotK0.east) -- node [below left] {$d_k^{r,i}$} (1,4) node [circle,draw,solid,minimum width=2mm,fill=cyan!60,thick] (land1) {} node [xshift=-0.1cm,left] {$z^i_k$};
		\draw [thick] (land1) -- node [above right] {$\hat{d}_{k+1}^{r,i}$} (robotK1.east);
		\draw [thick] (robotK1.east) -- node [right] {$d_{k+1}^{r,i}$} (6,4) node [circle,draw,solid,minimum width=2mm,fill=cyan!60,thick] (land2) {} node [xshift=0.1cm,right] {$z^i_{k+1}$};	
		\coordinate (robotK0east) at (robotK0.east);
		\coordinate (robotK1east) at (robotK1.east);
		\pic [draw, "$\alpha_{k+1}^i$", angle eccentricity=1.8,angle radius=1cm,angle radius=0.5cm] {angle = robotK1east--robotK0east--land1};
		\node at (robotK0) [below,yshift=-0.3cm] {$r_k$};
		\node at (robotK1) [below,yshift=-0.3cm] {$r_{k+1}$};
	\end{tikzpicture}
	\caption{Scenario showcasing a landmark moves in a circular path, where the proposed method fails to detect its dynamic nature.}
	\label{fig:FalseAssumption}
\end{figure}
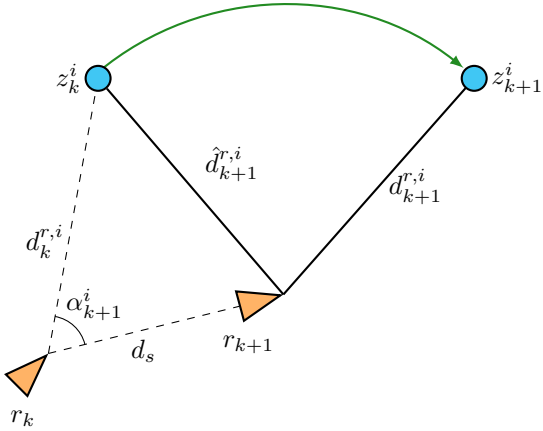

However, our method has a fundamental limitation: if a landmark rounds in a circle with a radius of $\hat{d}_{{k+1}}^{r,i}\pm \varepsilon$, we might misidentify it as stationary. While such a scenario is theoretically possible, it is scarce in real-world applications. This specific case is illustrated in Figure \ref{fig:FalseAssumption}, where our technique fails to correctly categorize the landmark's dynamic nature.

In Algorithm \ref{alg:ProposedMethod} you can see the proposed method for identifying the moving landmarks.
\begin{algorithm}
	\caption{Proposed moving landmark Identification}\label{alg:ProposedMethod}
	\For {each step $k+1$}{
		\For{each landmark $i$ in FOV}{
			Measure the distance between robot and landmark $i$ ($d_{k+1}^{r,i}$)	\\
			Compute the estimated stationary assumption distance between robot and landmark $i$ from step $k$: $\hat{d}_{k+1}^{r,i} = \sqrt{d_s^2 + \left(d_{k}^{r,i}\right)^2 - 2d_s d_{k}^{r,i}\cos(\alpha)}$ \\
			\If{$\left\lvert \hat{d}_{k+1}^{r,i} - {d}_{k+1}^{r,i}\right\rvert \leq \varepsilon$}{
				Add landmark $i$ to the state vector \\	
			}
			Run UKF-SLAM algorithm from Algorithm \ref{alg:UKFSLAM}
		}
	}
\end{algorithm}

\section{Experiments and Results}\label{sec:sim}
To assess the effectiveness of the UKF-SLAM algorithm in dynamic environments, we ran a series of MATLAB simulations that carefully varied specific parameters while considering real-world parameters. The main objectives consisted of:
\begin{enumerate}
	\item Evaluate adaptability in dynamic situations
	\item Evaluate the accuracy of robot position localization and the corresponding landmark geographical coordinates
\end{enumerate}

In our UKF-SLAM approach for moving landmarks, each moving landmark follows a generated random path with fixed velocity. We employ a method that ensures the path remains near the ground truth, thus keeping the landmark within the robot's field of view for an appropriate duration. This strategy guarantees landmarks do not drift far from the robot, facilitating better observation. Our evaluation systematically examined the algorithm's effectiveness across diverse scenarios by modifying three fundamental parameters in a structured manner: total waypoints ($M$), total landmarks ($N$), and moving landmarks $N_d$. Each scenario is differentiated by varying a single parameter while holding the others constant.
\subsection{Datasets}
This study focuses on establishing a simulation framework designed to generate datasets comprising waypoints, landmarks, and trajectories. The simulation allows users to specify the desired quantity of waypoints and landmarks, positioning them appropriately within a two-dimensional plane. To construct a continuous route, trajectories connecting these waypoints are computed using the Traveling Salesman Problem (TSP). These waypoint trajectories serve as ground truth data for subsequent experiments and facilitate the comparison between conventional and proposed algorithms.
A predetermined parameter determines the placement of landmarks, ensuring that each landmark is within the radius of the waypoints. The simulation uses an iterative process to ensure that the landmark compliance criterion is met. Landmarks that are initially eliminated are dynamically regenerated until they meet the requirenments. The outcome is a comprehensive dataset that offers diverse scenarios for SLAM experimentation. 
\subsection{Evaluation Metrics}
\begin{figure*}
	\centering
	\subfigure[Proposed Method Error\label{fig:ErrorPro}]{\includegraphics[width=0.45\linewidth]{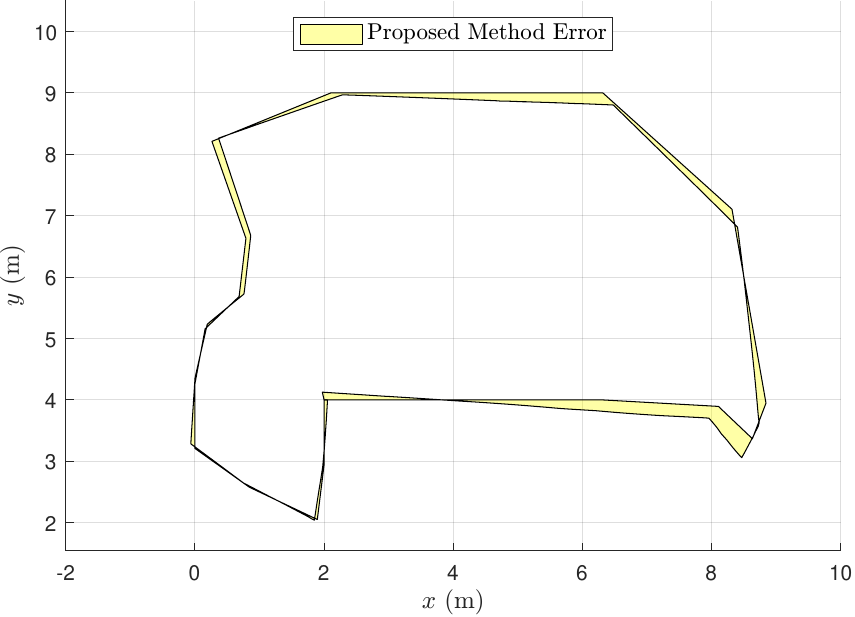}} ~
	\subfigure[Conventional Method Error\label{fig:ErrorConv}]{\includegraphics[width=0.45\linewidth]{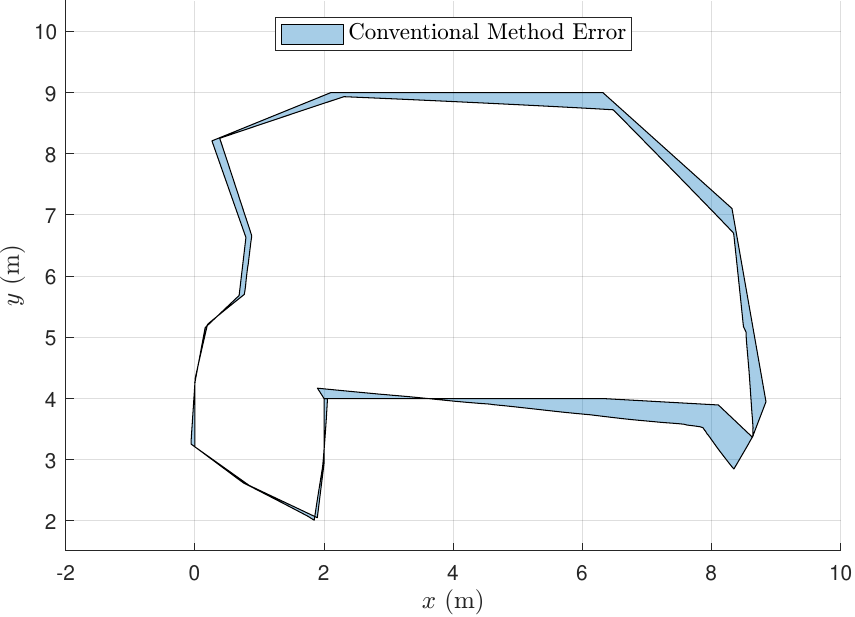}}
	\caption{Error of two algorithms in a single trial ($M= 20$, $N = 10$, and $N_d = 3$)}
	\label{fig:Error}
\end{figure*}

Valid evaluation metrics are essential for comparing proposed algorithms with conventional algorithms in different scenarios. In SLAM evaluations, it is essential to compare the estimated trajectory step-by-step with the ground truth. However, measurement errors can occur due to the lack of an exact one-to-one correspondence between the ground truth and the estimated points. To address this, we employ the Integral Absolute Error (IAE) \cite{Coughanowr2008wb} in our work as a criterion for measuring the error and the validity of our approach. Figure \ref{fig:Error} represents IAE metric for one experiment. The colored areas represent the error between the estimated path and the ground truth. The area of colored regions indicates the error.

\subsection{Effect of the Number of Moving Landmarks}
\begin{figure}
	\centering
	\includegraphics[width=\linewidth]{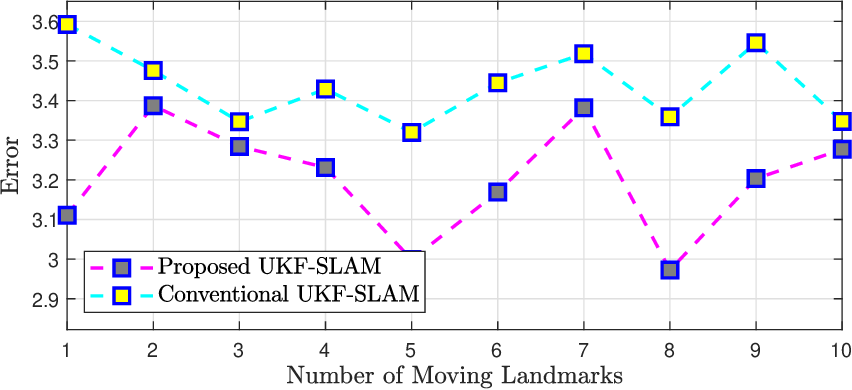}
	\caption{Error over changing the number of moving landmarks ($M = 20$ and $N = 10$)}
	\label{fig:movinglandmarks}
\end{figure}

In the first experiment, a consistent total of ten landmarks was maintained, with varying ratios of moving to stationary landmarks in each trial. The trial outcomes were determined by repeating each trial one hundred times and subsequently computing the mean error across all trial repetitions. As illustrated in Figure \ref{fig:movinglandmarks}, the results deviated from the expected linear trend, where the error typically increased with a higher proportion of moving landmarks. This deviation suggests a strong dependency on noise levels. It is important to note that noise was independently applied in each test, rendering the results of each test statistically independent from one another. The reason for the deviation from the expected linear trend is likely because noise can have a significant impact on the accuracy of the SLAM algorithm. When there are more moving landmarks, the noise is more likely to cause the algorithm to misidentify a landmark as moving, which can lead to an increase in error.

Nonetheless, the obtained results consistently demonstrate the superior performance of the proposed algorithm compared to the conventional one across all scenarios. The proposed algorithm achieved a lower error than the conventional algorithm in all experiments, even in the experiment with a high proportion of moving landmarks.

\subsection{Effect of the Number of stationary landmarks}
\begin{figure}
	\centering
	\includegraphics[width=\linewidth]{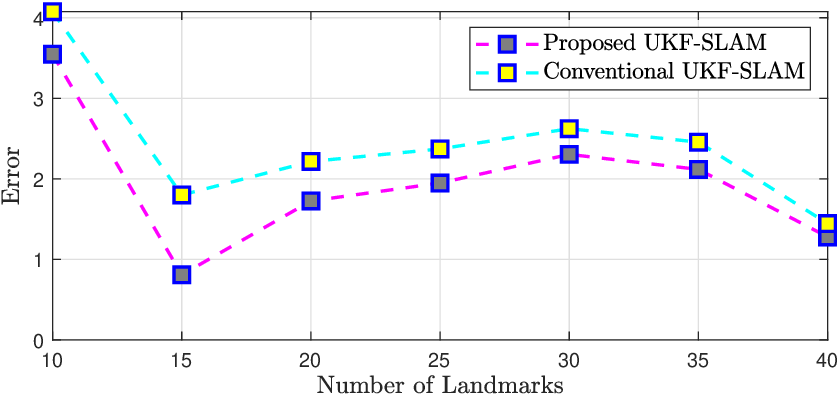} 
	\caption{Error over changing the total number landmarks ($M = 20$ and $N_d = 3$)}
	\label{fig:landmarks}
\end{figure}

In the second experiment, the number of movable landmarks was fixed at 3, and the number of stationary landmarks varied. The trial outcomes were determined in the same way as in the first experiment. As shown in Figure \ref{fig:landmarks}, the results showed that the error rate of the proposed algorithm consistently outperformed that of the conventional method across all measurements. This suggests that the proposed algorithm is more robust to the number of landmarks than the conventional algorithm.

The convergence of the results of the two algorithms as the number of landmarks increases can be attributed to the diminishing impact of the removed landmarks on the overall performance as the number of stationary landmarks grows. This is because the stationary landmarks provide a more stable reference for the SLAM algorithm, which can help compensate for errors caused by the moving landmarks.

The results of the second experiment further support the findings of the first experiment. They suggest that the proposed algorithm is a promising approach for SLAM in dynamic environments. The proposed algorithm is more robust to the number of landmarks than the conventional algorithm and can achieve lower error rates.

\subsection{Effect of the Number of Waypoints and Trajectory}
\begin{figure}
	\centering
	\includegraphics[width=\linewidth]{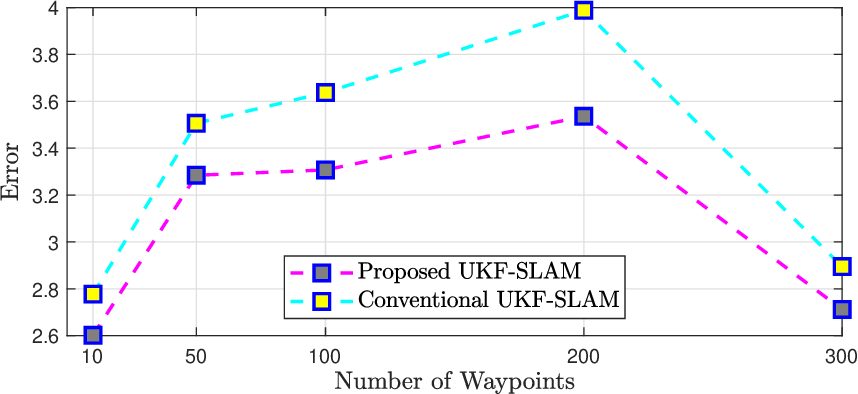}
	\caption{Error over changing the number of waypoints ($N = 10$ and $N_d = 3$)}
	\label{fig:Waypoints}
\end{figure}

\begin{figure}
	\centering
	\includegraphics[width=\linewidth]{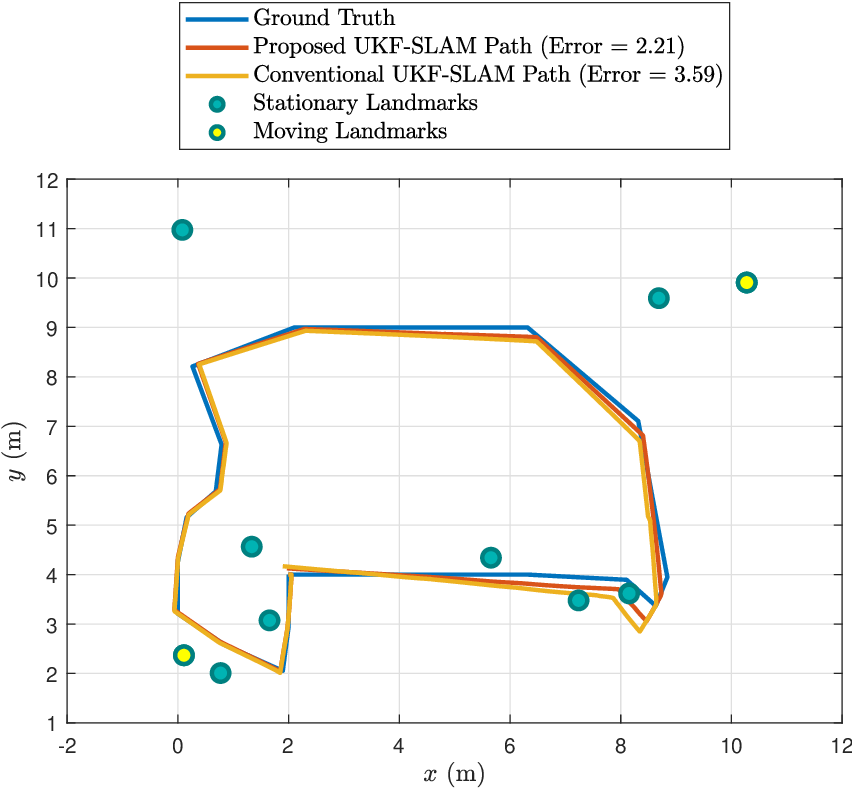}
	\caption{Comparative Analysis of Error Performance in Two Algorithms Applied to a Common Trajectory ($M = 20$, $N = 10$, and $N_d = 3$)}
	\label{fig:Map}
\end{figure}

In the third experiment, we varied the number of waypoints to measure the error affected by different trajectories. The error of the conventional algorithm was consistently higher than that of the proposed algorithm. This is because the proposed algorithm uses a more robust estimation method that is less sensitive to the arrangement of the landmarks in the environment. We repeated each trial a hundred times, and although the number of waypoints was the same for each trial, the trajectories of each trial were different. However, all parameters were the same in both algorithms in each trial. The mean of the trails is shown in Figure \ref{fig:Waypoints}. Figure \ref{fig:Map} shows a trail for both algorithms with three moving and seven stationary landmarks and 20 waypoints.

Both algorithms indicated a positive correlation between reduced step length and improved performance. This is because a smaller step length results in a smoother trajectory, which is easier for the algorithm to estimate.

Figure \ref{fig:Map} presents a comparison of the error between two algorithms employing identical parameters and trajectories. The error between the estimated trajectory and the ground truth is represented by the colored regions between them.

\subsection{Timing and Computational Efficacy}
\begin{figure}
	\centering
	\includegraphics[width=\linewidth]{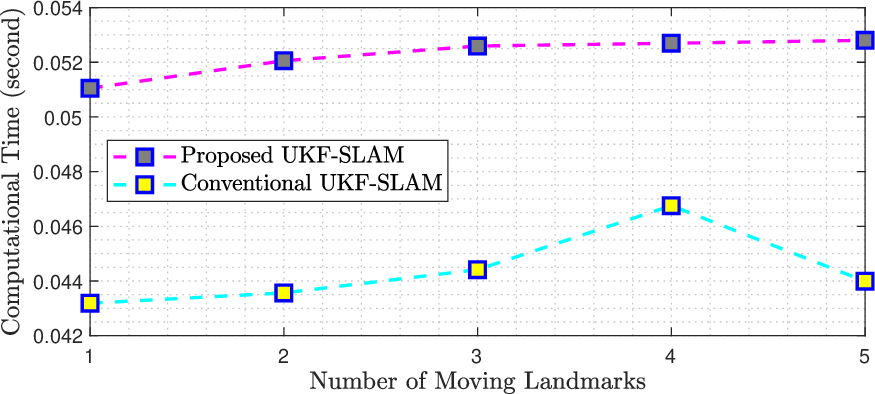}
	\caption{Computational time vs. the number of moving landmarks ($M = 20$ and $N = 10$}
	\label{fig:Time}
\end{figure}

In the final experiment, we evaluated the implementation time of the proposed algorithm and the conventional algorithm. We kept the total number of landmarks fixed at ten and the number of waypoints the same in all trials. We repeated each trial a hundred times and recorded the mean error. The dynamic processing block took some time to implement. However, reducing the size of the state vector saved time in the processing unit. Overall, the implementation time for both algorithms leveled out at around 0.05 milliseconds (ms) and 0.04 ms for the proposed and conventional algorithms, respectively. Figure \ref{fig:Time} compares the results of the two algorithms.

\subsection{Real-World Scenario Mapping}
\begin{figure*}
	\centering
	\subfigure[IUT Stochastic Signal Processing Lab - 2D Map\label{fig:IUTRad}]{\includegraphics[width=0.45\linewidth]{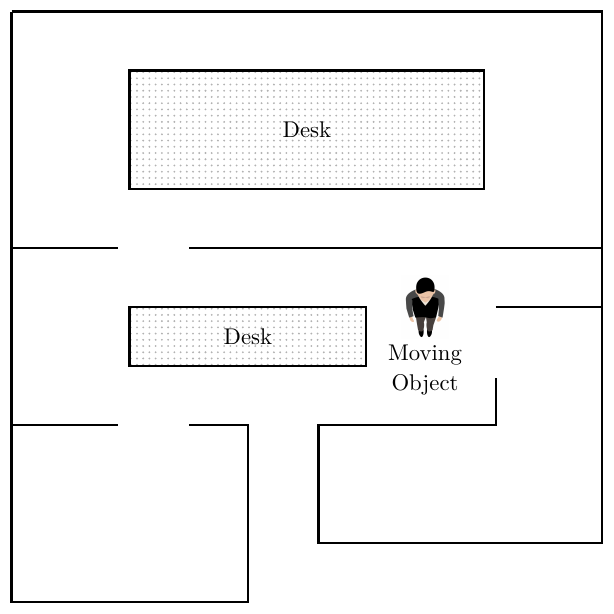}} ~
	\subfigure[Constructed Map\label{fig:ConstMap}]{\includegraphics[width=0.45\linewidth]{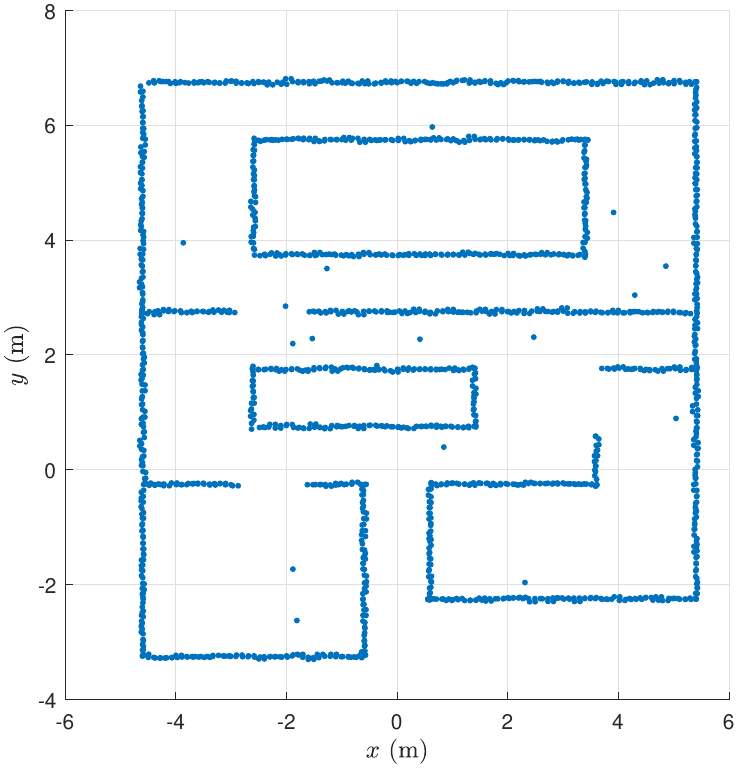}}
	\caption{Real-wold scenario mapping with a single moving object.}
	\label{fig:IUTMapping}
\end{figure*}

This experiment aims to replicate robot movement in an indoor environment and analyze its realistic application. Unlike previous experiments, which used dispersed landmarks as distinct points with specified distances, this study considers a more authentic simulation where each robot observation is regarded as a landmark. For the MATLAB simulation, these landmarks were approximated using a large number of closely situated points, which essentially served as the observed landmarks, representing the robot's observation at each movement step. Figure \ref{fig:IUTRad} shows the true representation of the simulation environment, while Figure \ref{fig:IUTMapping} displays the mapping algorithm's output, which features a single moving object.

\section{Conclusion}\label{sec:con}
In this paper, we proposed a method for eliminating moving landmarks from a SLAM system using Euclidean distance measurement. The method predicts the robot's position in the next step and calculates the distance between the robot and each landmark. If there is a significant difference between the calculated and measured distances, it identifies that as a moving landmark to eliminate further processing. This process improves algorithm performance in dynamic environments by ensuring only reliable stationary landmarks are considered. The enhanced UKF-SLAM algorithm increases adaptability to real-world scenarios with prevalent environmental dynamics while improving SLAM output quality.

Using a simulated dataset, the effectiveness of the proposed technique for moving landmark removal was evaluated and demonstrated. The proposed method has several advantages over existing methods. It is simple to implement and does not require any prior knowledge of the environment. Moreover, our method was robust in the presence of noise and variations in the robot's motion. To thoroughly evaluate our methodology, we conducted four primary experiments comparing the results of the proposed algorithm to those of a conventional algorithm. While the proposed algorithm required more computational time, it outperformed the conventional approach in other performance metrics. Notable is that both algorithms were affected by real-world noise, which we incorporated on purpose into our simulations to increase their realism. To reduce noise effects and ensure equal algorithmic comparisons, we repeated each experiment a hundred times and plotted the averaged results.

The proposed algorithm is a promising step towards developing more robust and accurate SLAM systems for dynamic environments. While further improvements are needed for real-world applications, we have conducted simulations to thoroughly evaluate the effectiveness of our proposed algorithms in real-world scenario mapping. This rigorous approach allows us to assess our methods in practical conditions and gain insights for future enhancements.

Our work introduces a new filter-based SLAM algorithm that uses the UKF and a dataset designed specifically for dynamic environments. This research highlights the effectiveness of filter-based SLAM methods in dynamic scenarios, an area often overlooked in SLAM research. Further exploration of filter-based SLAM in dynamic environments is encouraged to redefine the field. In conclusion, the enhanced UKF-SLAM algorithm demonstrates the efficacy of filter-based SLAM algorithms in dynamic environments and opens up new possibilities for improving the accuracy and robustness of SLAM systems. 

\bibliographystyle{ieeetr}
\bibliography{mybib.bib}

\vfill

\end{document}